%% file: main.tex
\documentclass[sigplan,screen]{acmart}                

\copyrightyear{2024}
\acmYear{2024}
\setcopyright{acmlicensed}\acmConference[FDG 2024]{Foundations of Digital Games 2024}{May 21--24, 2024}{Worcester, Massachusetts, USA}
\acmBooktitle{Foundations of Digital Games 2024 (FDG 2024), May 21--24, 2024, Worcester, Massachusetts, USA}
\acmPrice{15.00}
\acmDOI{10.1145/3582437.3587208}
\acmISBN{978-1-4503-9855-8/23/04}

\AtBeginDocument{%
  \providecommand\BibTeX{{%
    \normalfont B\kern-0.5em{\scshape i\kern-0.25em b}\kern-0.8em\TeX}}}

\usepackage{todonotes}
\usepackage{multirow}
\usepackage{subcaption}
\usepackage{colortbl}

\usepackage{graphicx}

\begin{document}

\title{The Ink Splotch Effect:\\A Case Study on ChatGPT as a Co-Creative Game Designer} 


\author{Asad Anjum}
\email{asad.anjum@nyu.edu}
\affiliation{%
  \institution{New York University}
  \city{Brooklyn}
  \state{New York}
  \country{USA}
}

\author{Yuting Li}
\email{yl7685@nyu.edu}
\affiliation{%
  \institution{New York University}
  \city{Brooklyn}
  \state{New York}
  \country{USA}
}

\author{Noelle Law}
\email{noelle.t.law@gmail.com}
\affiliation{%
  \institution{Independent Researcher}
  \city{Baltimore}
  \state{Maryland}
  \country{USA}
}

\author{M Charity}
\email{mlc761@nyu.edu}
\affiliation{%
  \institution{New York University}
  \city{Brooklyn}
  \state{New York}
  \country{USA}
}

\author{Julian Togelius}
\email{julian@togelius.com}
\affiliation{%
  \institution{New York University}
  \city{Brooklyn}
  \state{New York}
  \country{USA}
}


\renewcommand{\shortauthors}{Anonymous}

\begin{abstract}
This paper studies how large language models (LLMs) can act as effective, high-level creative collaborators and ``muses'' for game design. We model the design of this study after the exercises artists use by looking at amorphous ink splotches for creative inspiration. Our goal is to determine whether AI-assistance can improve, hinder, or provide an alternative quality to games when compared to the creative intents implemented by human designers. The capabilities of LLMs as game designers are stress tested by placing it at the forefront of the decision making process. Three prototype games are designed across 3 different genres: (1) a minimalist base game, (2) a game with features and game feel elements added by a human game designer, and (3) a game with features and feel elements directly implemented from prompted outputs of the LLM, ChatGPT. A user study was conducted and participants were asked to blindly evaluate the quality and their preference of these games. We discuss both the development process of communicating creative intent to an AI chatbot and the synthesized open feedback of the participants. We use this data to determine both the benefits and shortcomings of AI in a more design-centric role.


\end{abstract}

\keywords{LLMs, AI-assisted game design, co-creative game design, mixed-initiative creativity, procedural content generation, user study, Unity}

\maketitle


\section{Introduction}

Despite the rising prominence of artificial intelligence (AI) in both research and retail products, it remains a sensitive and even taboo topic in creative circles\footnote{https://www.cnn.com/2022/10/21/tech/artists-ai-images/index.html}\footnote{https://www.forbes.com/sites/joemckendrick/2022/12/21/who-ultimately-owns-content-generated-by-chatgpt-and-other-ai-platforms/?sh=891092b5423a}\footnote{https://www.ign.com/articles/dragon-age-writer-dismisses-ai-generated-storytelling-bioware-david-gaider}. The surge of AI art has served as a further catalyst to this issue, and perhaps no other field embodies the fear and rejection of newer AI principles than game development\footnote{https://www.cbc.ca/radio/video-games-artificial-intelligence-1.6974408}\footnote{https://fortune.com/2023/07/25/video-game-studios-scared-ai-forcing-managers-study-machine-learning-offering-employees-7000-bounties-gala-sports/}\footnote{https://www.gameinformer.com/2023/11/22/its-undeniably-going-to-cost-people-jobs-inside-the-game-industrys-fight-over-ai}. It is a creative discipline comprised of artists, composers, designers, writers, actors, and programmers, all of which are feeling the reverberate effects of AI advancements. It is unsurprising, then, that a large portion of this industry is skeptical\textemdash even fearful\textemdash of the effect that AI can have.

\begin{figure}[ht!]
    \centering
    \includegraphics[width=0.75\linewidth]{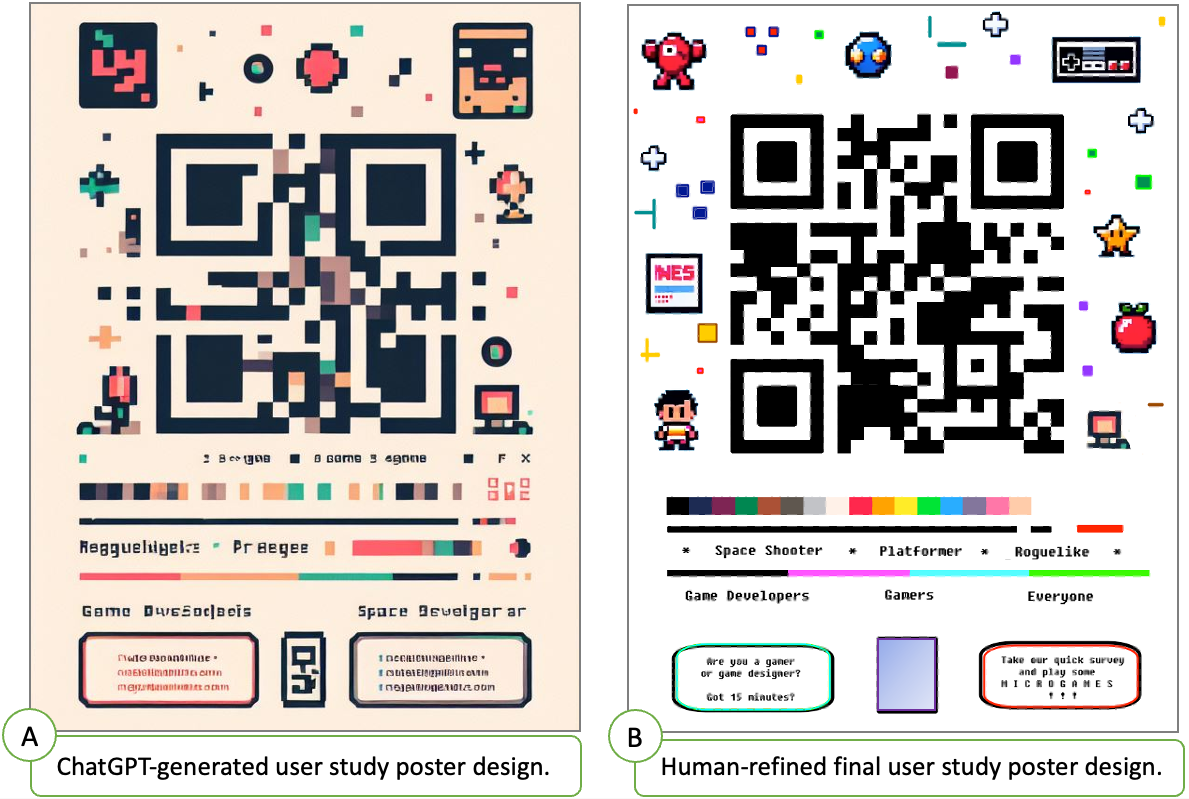}
    \caption{\textbf{Leveraging LLMs for creative tasks can be highly efficient, however there still remains a need for human-refinement.} Much like this poster, where a human refined the AI-generated art into something readable and more aesthetically pleasing, humans can use LLMs as a `muse' to design better games. } 
    \label{fig:poster}
\end{figure}

What becomes important then, not just to the advancement of AI but to its acceptance, is drawing a clear distinction between what AI tools can provide and what still requires a human-refinement. If AI is simply a tool used to improve and expedite certain processes in creative fields, it should be inferior at making cohesive, expressive creative choices compared to a designer with a clear vision for a project. Moreover, if an assistant AI tool is given full autonomy over what decisions to make and how to implement them in a creative medium such as game design, the result should theoretically be functional and structurally sound, but lack creative vision. Consequently, this idea was put to the test\textemdash could a general AI tool such as OpenAI's ChatGPT create a game with genuine, well-defined creative vision with little to no influence from the user? If so, how would those creative choices compare to ones made by a designer instead? 

This paper attempts to answer these questions by having game designers create a handful of common game archetypes with minimalist gameplay. We then compare their game feel, uniqueness, and cohesion to the same games made only using suggestions and code from LLMs. We asked participants to evaluate the quality, game feel, innovation, and overall preference of three games \textemdash a base game, a human-designed game, and an AI-suggested game\textemdash across three genres using a ranking scale questionnaire. The results of the questionnaire illuminated the strengths and weaknesses of both methods of pure human design versus AI-assisted design, and how a collaboration between human game designers and a general AI assistant could be beneficial or detrimental to the creative process.

\section{Background and Related Works}

\subsection{Collaborative Game Design}
Collaborative systems allow AI agents to interact in the development process with the game designer by generating or modifying some aspect of the game. In past works, these systems have typically been applied to level design \cite{alvarez2018fostering} \cite{bhaumik2021lode} or for developing entire games \cite{nelson2008interactive}\cite{cook2016angelina}. Computational creativity from these collaborative systems can be applied to any part of the game, such as level design, visuals, narrative, or gameplay \cite{liapis2014computational}. The extent of interactivity of the agent can come in varying degrees of involvement; the AI can act as a guiding system, as an equal collaborator, or can simply model user intent \cite{guzdial2019friend}. The design process for these collaborative systems \cite{partlan2021design} and how they are used by their designers\textemdash whether by directly incorporating their content into the game \cite{vimpari2023adapt} or as a source of inspiration akin to the artistic exercise of Bulletism \cite{Artsapien_2023} \cite{creative_future_gdc}\textemdash has become an active area of research. The benefits and drawbacks of these collaborative AI systems are critical to explore from a human designer and a researcher perspective, particularly as generative systems (e.g. text-to-image, LLMs) become more widely used. We use ChatGPT as an abstract AI-assisted content generator\textemdash one intended for the use of designing game feature and game feel content.

 

\subsection{Designer Intent}

For creative projects such as video games, the designer's intent is expressed thoroughly through the different aspects of the game. This expression can take form in the game's art, music, level design, narrative, and general gameplay\textemdash all of which contribute to the overall player experience and feel of the game. Automating game design while maintaining original game designer intent is a challenge, especially when given limited context and understanding on the AI's part. Nelson et. al. first explored how the communication and discrepancies between AI game design assistants and human users affect the quality and experience of a game \cite{nelson2008interactive}. This was later expanded by Treanor et. al. to build entire games from user `ideas' in a mixed-initiative process \cite{treanor2012game}. Various other works have looked to model and expand the creative intent of the designer and apply the output to generative game content \cite{liapis2013designer} \cite{hjaltason2015game} \cite{pichlmair2021designing} or games themselves \cite{togelius2008experiment} and then evaluate them on the player experience. In this experiment, we use ChatGPT to examine how well an LLM can understand and contribute to the game development process based on a human designer's original intent. We prevent the LLM from being involved directly with the output product, and rather implement the game features and code suggestions through chat communication. 


\subsection{LLM-Assisted Game Content Generation}

The development and growing accessibility of large language models (LLMs) for the use of content generation has become increasingly relevant in game design. Most of the use cases of LLMs in PCG\textemdash which use the iterations of the GPT and LLaMA models\textemdash are used to generate game assets and more abstract content such as suggesting game features \cite{charity2023preliminary}. Concretely, LLMs have been used to generate levels for previously researched PCG domains such as Sokoban \cite{todd2023level} and Mario \cite{sudhakaran2023prompt}. LLMs have also been used for narrative generation to facilitate story telling \cite{sun2023language} and quest generation \cite{vartinen2022generating}. For data augmentation, these models can be applied to text-to-asset input prompts to train other generative models such as for sprites and levels \cite{merino2023five}.

Unlike previous works, where the output of the LLMs pertained to a single aspect of the game, we apply the output of the LLM in the context of the entire game design process. With this system, we use ChatGPT's GPT-3.5 model to suggest game features, suggest game feel elements, and to generate code in the C\# language for the Unity game engine. We combine both the abstract and concrete content generation capabilities and apply them directly to the game design process. 


\begin{figure*}[ht!]
    \centering
    \includegraphics[width=\linewidth]{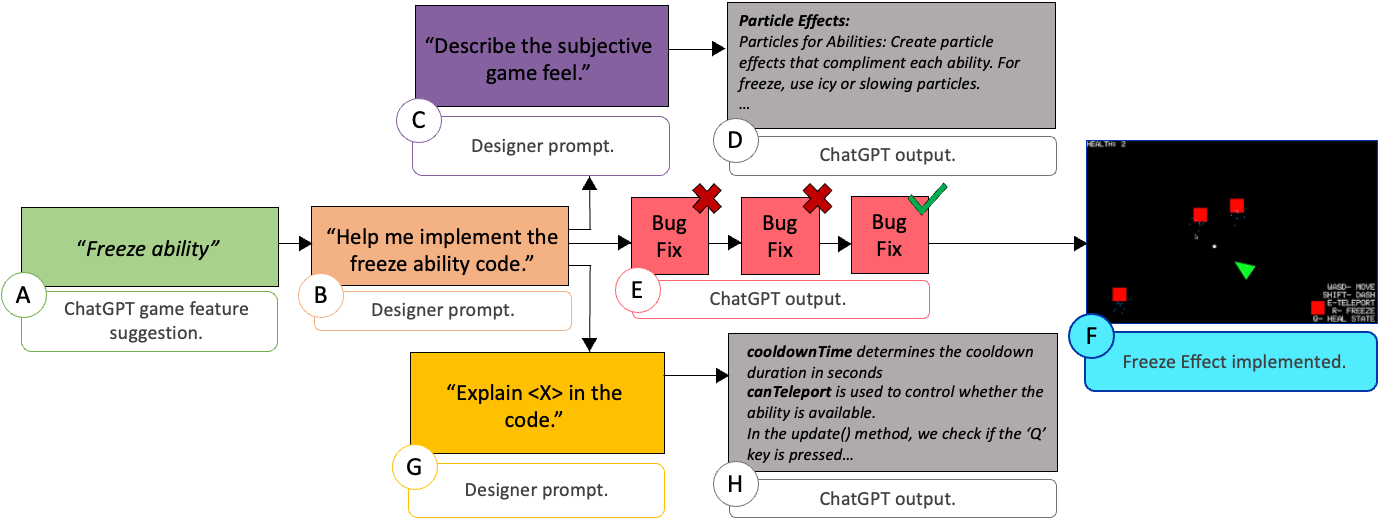}
    \caption{\textbf{Example of the freeze ability being implemented for the space shooter genre game with guidance from ChatGPT.} The freeze ability is (A) first suggested by ChatGPT from an initial prompt. (B) The user then asks ChatGPT for help with implementing the code in Unity engine in C\#, (E) the output code from ChatGPT is attempted to be implemented 3 times by the user until it is successfully integrated into the game without error. The user also asks for (G) explanations about the generated code and (C) further suggestions to add game feel.} 
    \label{fig:ex-feat}
\end{figure*}


\section{Methodology}

To test the suggestive game design capabilities of ChatGPT, a series of 2D games for three game genres \textemdash arcade space shooter, platformer, and turn-based roguelike\textemdash were designed. Each genre had three games: a baseline game; a pure human-designed game; and a game made from ChatGPT suggestions The games were designed with minimalist graphics, where players and NPCs were made up of simple shapes and colors, to prevent any visual bias across genres.

These genres were chosen to test ChatGPT's capabilities for general game domain knowledge and how well it could apply context-relevant design suggestions. The platformer was chosen due to its popularity for Unity game design tutorials. The space shooter was chosen to represent a classic genre for arcade games. The roguelike was chosen as a genre of game with the most complex mechanics (i.e. procedurally generated maps and AI behaviors) and a different time signature (being turn-based). These selections attenuated potential bias in the user study from participants who had a preferential genre of game to play. 

The AI trials mimicked a situation where a developer had a basic gameplay loop in mind. These designers would then turn to ChatGPT for a comprehensive tutorial of how to implement their idea and what mechanics or abilities could be added to improve gameplay. We assume for these experiments that the developer had blind faith in the ideas presented to them, and if a range of items or ideas were ever provided, they would simply ask which of the options was most recommended. 
ChatGPT was prompted for everything from high level subjective questions (e.g. ``What abilities should I add to my game?'') to low level code implementations (e.g. ``Can you provide code and instructions on how to apply this in Unity?'') to add features and improve the game feel of each game. Author Steve Swink describes "game feel" as "the tactile, kinesthetic sense of manipulating a virtual object" and "the sensation of control in a game" \cite{swink2008game}. 
The authors performed the role of novice human designers for this study and directly implemented ChatGPT's responses (without regard to personal opinion of the output) into the games.
In addition, improvements to game feel and general bug fixes were limited to suggestions exclusively from ChatGPT. A number of limitations were placed on prompting ChatGPT to ensure the prompter’s bias or influence was kept to a minimum. The prompter was only allowed to request code implementation for up to 5 different features for each game. If a bug arose from any of the generated code, up to 3 attempts could be made to ask the program how to fix the error, after which its failure to do so would be noted. The prompter could request 1 general explanation of the provided code and could ask once which game feel elements should be associated with the feature.

The following subsections describe the distinguishing features and creative development process from implementing the 9 games. 

\subsection{Space Shooter}

\begin{figure}[h!]
    \centering
    \includegraphics[width=\linewidth]{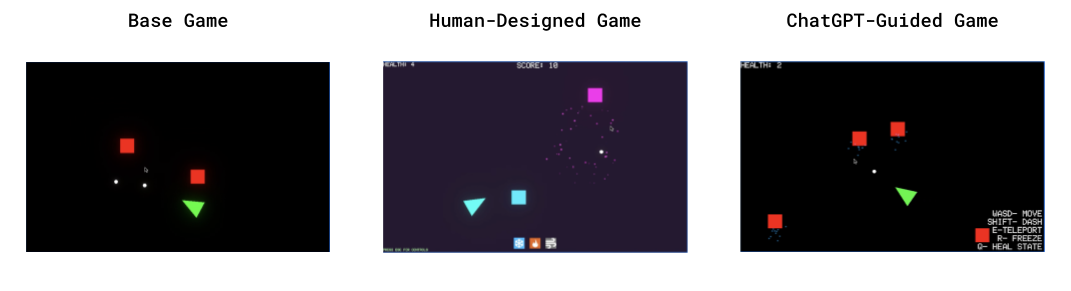}
    \caption{\textbf{Screen-captures of the 3 shooter games created for the experiment.}}
    \label{fig:shooter-games}
\end{figure}

The prompt for the first game was to "Create a fast paced, top-down shooter with 3 special abilities. Each ability should have a short cooldown. The goal is to survive for as long as possible while enemies randomly spawn into the scene and chase the player".

\subsubsection{Base Game} The base shooter game included standard top-down movement as well as a short ranged omni-directional dash on a brief cooldown. The player would be oriented towards the mouse position on screen, and a button press would fire a bullet in the same direction. Pressing and holding the mouse button would fire bullets at a constant rate. Enemies would spawn into the edge of the scene at set intervals, from any one of 9 evenly spaced spawn points. Enemies would follow the player's position at a constant speed and destroyed when they either collided with the player, in which case the player's health would drop, or shot by a bullet. The game featured a black background and simple geometric shapes for the player, bullets, and enemies, with no particle effects or generally accepted elements of game feel. The only user interface (UI) element was a countdown on the top left indicating the player's current health.

\subsubsection{Human-Designed Game}

The initial design process began with adding general game feel elements to the base game. This included shaking the screen any time an enemy was hit, adding appropriate sound effects for shooting and destroying enemies, and triggering an explosive particle effect when enemies were killed. During playtesting, we felt that taking damage was not effectively communicated to the player, despite the health counter updating on the UI. We resolved this by incorporating a short, screen-wide glitch effect during enemy-player collision. The more explicit visual cue influenced future decisions regarding visual design and ramping difficulty, and its clarity in frantic scenarios supported a more chaotic gameplay loop. This constant re-assessment of the game’s direction through multiple iterations of testing and refinement was vital to the human design process. It presented an important factor in delineating between human and AI driven games: the ability to ask \textit{why}.

A screen-wide freeze ability\textemdash which would stop all enemies currently on screen in an instant\textemdash was first implemented into the game. In its original state, enemies would suddenly stop moving when the associated key was pressed, offering little to no feedback to the player. As the first of many changes to rectify this, a strong screen shake effect was added to signify that a powerful attack had been activated. Frozen enemies would also turn light blue as opposed to their standard red, a faded blue vignette would appear along the edges of the screen accompanied by the sound of ice cracking and forming, and destroying frozen enemies would trigger a new "shattered" sound effect. Each of these additions were intended to give the ability a distinctly ``cold'' and elemental nature.

Once this ability had been set, it felt natural to continue along the “elemental” theme for the game. A sweeping fire attack\textemdash which destroyed enemies on impact\textemdash  and a strong outwards wind blast\textemdash which pushed all enemies away from the player\textemdash were implemented to complement the freeze ability. Both were given their own screen-wide colored vignettes to match the freeze effect (orange for fire, white for wind) as well as appropriate sound effects and specialized particle effects. The wind ability pushed white lines from the center outwards, and the fire ability produced an arc of flames around the player.  

With the game mostly complete, playtesting and refinement were once again responsible for many small but important changes to give the game its final look and feel. Cooldowns for abilities were visually indicated on the UI and each skill had a different cooldown time based on how powerful the ability was. To add to the slightly frenetic nature of the game, a slight amount of screen shake was invoked every time the player fired a bullet. A faded red vignette would appear when the player was hit, and the colors of both the background and the enemies would change upon taking damage.

\subsubsection{ChatGPT Guided Game}
The initial task for the ChatGPT guided game was to recreate the entirety of the base game, noting any failures or differences in implementation from the original. It was able to accurately rebuild most the game from scratch, with detailed descriptions of how to add objects, code, and components such as line renderers in Unity. In some instances, it actually employed better practices than the base game, including setting up a more robust enemy spawning system. However, the program did run into roadblocks. For instance, it failed to correctly diagnose why the player's rotation did not exactly match the mouse position. While the process was fairly harmonious, proper execution required a more in-depth understanding of Unity and the general game making process.

Once the base game was set, we asked ChatGPT to suggest a list of potential abilities for a "top-down, spaceship shooter game". It provided a list of 20 options, including:

\begin{itemize}
    \item \textbf{Teleportation}: Instantly jump to a different location on the screen, useful for escaping dangerous situations or repositioning strategically.
    \item \textbf{Nanobot Repair}: Activate a healing ability that slowly restores your ship's health over time.
    \item \textbf{Time Manipulation}: Slow down or temporarily freeze time to dodge bullets, aim precisely, or gain a tactical advantage.
\end{itemize}

While the code for each ability was provided on request, a number of issues were met in the process. For instance, teleportation did not achieve the desired effect that ChatGPT itself outlined in its description, as it would often teleport the player directly to an enemy or to the same position they already were. A later issue with the same ability would cause the player to disappear every time it was activated. Despite multiple attempts to solve this problem, ChatGPT was not able to resolve the issue, and the problem was fixed manually by recognizing the player's location on the z-axis was being changed during teleportation.

The model struggled most with incorporating game feel elements for the abilities. It often gave vague instructions for adding particle and sound effects, contrary to ChatGPT's in-depth explanation on utilizing the line renderer component (used as an accessory to the dash function in the base game). Basic suggestions such as having the teleport look like "a burst of stars or particles that appear and disappear quickly" were provided, but without detailed instructions on how to achieve them. As a result, the final game contained a significantly lower count of additional game feel elements than the human-designed counterpart. The ChatGPT log for the space shooter genre game can be accessed here\footnote{https://chat.openai.com/share/f807763a-6a25-4fd3-85de-adb5f1de8159}.

\subsection{Platformer}

\begin{figure}[h!]
    \centering
    \includegraphics[width=\linewidth]{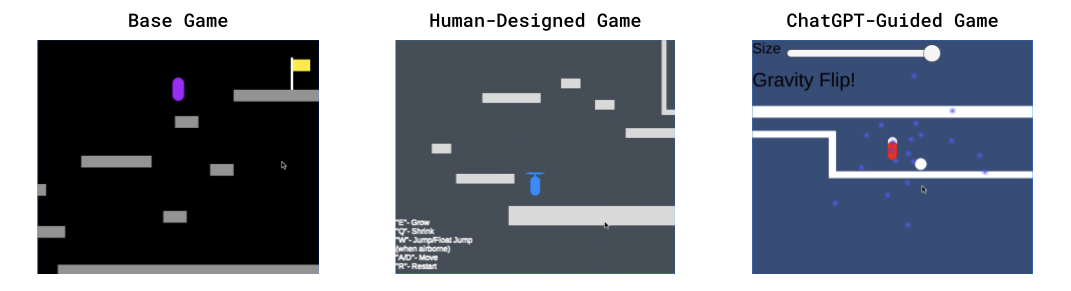}
    \caption{\textbf{Screen-captures of the 3 platformer games created for the experiment.}}
    \label{fig:platformer-games}
\end{figure}

The prompt for the second game was to ``Create a simple platformer where you need to reach the flag to win. Develop 3 transformative mechanics to the player which change their platforming. Design a short level that takes these abilities into account”.

\subsubsection{Base Game}
The base platformer game contains minimalist controls and visual design. It contains only horizontal movement, gravity, and a jumping mechanic to allow the player to traverse floating platforms found in the level. The player must reach the flag object found in the level to complete the game. The level is designed similar to that of a pyramid, where platforms are placed intermittently vertically and horizontally in the space. If the player falls off the level, they are respawned to the original starting point. Like the base space shooter game, the base platformer game contains no particle effects, sound effects, or any other elements of noticeable game feel. The only UI is a transparent screen that tells the player when they have won the game. 

\subsubsection{Human-Designed Game}
Once again, the design process began with implementing a layer of general game feel elements to the base game. The original jump function would sometimes cause the player to fall when attempting to jump at the edge of the platform. This was due to a lack of ``coyote jumping'', which describes the phenomenon of a player's ability to jump for a split second after moving off of the platform. An innate understanding of platforming design allowed responsiveness to be improved.

``Transformative mechanics” in the platformer prompt was interpreted by the designer as the player having shape-shifting abilities. Much like the elemental abilities in the shooter, this allowed for a level of coherence in potential abilities while sacrificing their range of options. Shrinking and growing abilities were both added, each with their own control inputs. When small, the player could traverse thin crevices, but suffered from a much smaller jump. When big, they would be unable to pass through certain environments, but were able to jump proportionally higher. The surrounding level was altered to encourage the use of these abilities, and the camera scaled accordingly to keep the player in frame. Particle effects were added to both transformations, with a clump of small particles moving towards the player during the “growing” animation and particles moving outwards during the “shrinking” animation. 

The final ability allowed the player to extend their jumps by descending at a slower rate than normal. This could only be activated while the player was in the air, and would instantiate a small object mimicking the shape of an umbrella or hang glider on top of the character model. A small effect was added to the top of the player, where particles would push out upwards and then drop down off to the side, as if the “hang glider” was protruding directly from the top of the player's head. Additional improvements such as linearly interpolating camera zoom for smoother movement rounded out the remainder of the game.

\subsubsection{ChatGPT-Guided Game}

While ChatGPT was once again able to recreate the base game eventually, its poor understanding of the game’s context became an even bigger issue for the platformer. It would often provide code for a 3D project- despite multiple reiterations that the game was in 2D- and was incapable of producing a reliable method of following the player with the camera. It also listed a time manipulation ability (similar to that found in the space shooter, which would pause all enemies on screen for a brief period of time) as one of its 3 suggestions for “transformative mechanics”. This was despite the game not featuring any enemies. As a result, only 2 abilities had any real functionality in the final game. 

The first of these abilities was size manipulation, which was notably similar to the first two abilities in the human-designed game. The implementation was generally seamless, albeit missing many additional effects present in the human version and only featuring two states: “normal” and “giant”. The second ability allowed the player to flip gravity, however this only worked in one direction when implemented exactly as described. This lead to another manual fix by the prompter, who recognized the need to flip the player character over the x-axis when on the roof for the code to operate correctly. 

Unique to the platformer game, ChatGPT was also asked to give suggestions for level design given the transformative mechanics at play. These suggestions were generally indeterminate, but occasionally relevant, such as advising that the game have a section which was too narrow for the player to fit through if they were in their “giant” state. It did not, however, suggest that the level should feature an area that was not traversable by the “normal” state, which meant the final game could be completed without using the size manipulation ability altogether. Game feel suggestions were once again minimal, apart from a UI element which displayed the player’s size through a slider at the top left of the screen. The ChatGPT log for the space shooter genre game can be accessed here\footnote{https://chat.openai.com/share/e1813d47-d568-475e-9f57-5975cc42fce0}.

\subsection{Roguelike}

\begin{figure}[h!]
    \centering
    \includegraphics[width=\linewidth]{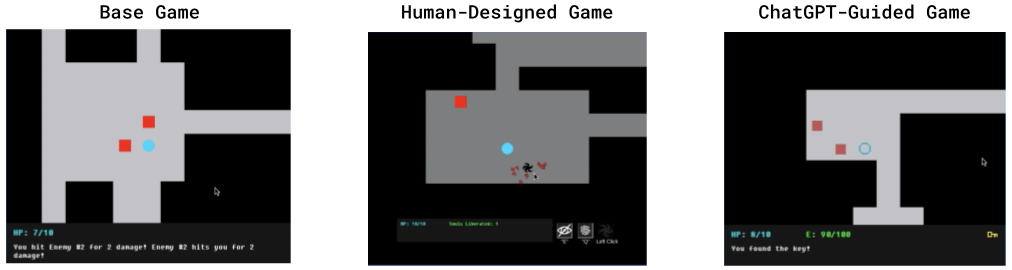}
    \caption{\textbf{Screen-captures of the 3 roguelike games created for the experiment.}}
    \label{fig:roguelike-games}
\end{figure}

The prompt for the roguelike game was to "Create a turn based stealth roguelike where the player must collect a key and then head to an exit to escape, all while avoiding patrolling enemies. The player can collect or recharge up to 3 abilities, but they must be stealth related. The player can also engage in direct combat with enemies".

\subsubsection{Base Game}
The base roguelike game takes elements from classic dungeon-crawler games. Rooms of a random size are procedurally generated and connected with corridors using a binary-space partition algorithm. The player's and enemies' movement and combat are turn-based, where each step the player enacts the next turn of the game. The player can move in any directional space (up,down,left,right) so long as there is a floor tile. The player can also skip a turn and remain on the same tile. To win the game, the player must find a randomly placed key, survive combat from enemies, and then use the key to unlock a randomly placed door. If the player loses too many health points, they lose the game. Enemies are randomly placed in the rooms, move in a random direction every 3 turns, and follow the player when they are within 5 spaces of them.  

\subsubsection{Human-Designed Game}
The roguelike separated itself from the previous two genres largely due to its turn-based nature, which impacted the approach to improving the game. As the first of 3 stealth mechanics, the player would become invisible for a number of turns. Enemies would lose sight of the player and would continue on random patrol paths, even if the player was in proximity. The ability was coupled with a vanishing sound effect and semi-translucent, cloud-like particle effect at the player's position. The player would then turn translucent themselves for as many turns as the effect held, after which they would turn opaque and attract enemy attention once again. The player was able to attack enemies in the invisible state, but this alerted the enemy to their position and instantly removed the invisibility effect.

The second ability\textemdash a smoke bomb\textemdash followed a similar structure, briefly limiting enemy sight lines and using a cloudy particle effect. An area around the player was enveloped by smoke, and all enemies inside the effect’s radius would completely stop. Unlike invisibility, enemies were not able to identify the players location in the smoke, even when the player was directly attacking them. This added benefit, along with the wider area of effect, made the smoke bomb significantly more effective than invisibility. This was offset by a longer cooldown. Finally, the player would be able to use shurikens for long-ranged combat. While the shuriken ability was available, a line renderer would indicate which direction the player was aiming.

Each ability was given distinct sound effects, and clashing sword sounds were added to the face-to-face combat from the base game to maximize feedback. Enemies would explode into a burst of blood when killed and cooldown timers were added to the user interface. To incentivize combat, killing enemies shortened all cooldown times by a sizeable amount.

\subsubsection{ChatGPT Guided Game}
Unlike the other 2 genres of games, ChatGPT was unable to recreate the original design of the base game for the roguelike genre. It could not understand the `turn-based' nature of the game and apply it to a real-time Unity game engine context. The LLM was also unable to correctly implement the binary-space partition algorithm for the rooms and the turn-based enemy AI movement. This could be due to the lack of context for the input prompts sent to the LLM, the Unity game engine, or the miscommunication between user intent and what ChatGPT was capable of creating. Therefore, the original base game was used as a starting point for implementing the feature suggestions of the LLM. The ChatGPT log for the roguelike genre game can be accessed here\footnote{https://chat.openai.com/share/32afedb3-6ef5-4578-81e7-a81874acdacf}. 

Like the human-designed version of the roguelike, ChatGPT suggested to implement a temporary invisibility feature, where the player could walk by enemies without engaging in combat or be attacked in retaliation. It also suggested to implement an energy system into the game. With this energy system, players would need a certain amount of energy points in order to engage in combat and turn invisible. This energy could be restored by waiting turns. In addition to the random chance to regain health from defeating an enemy, ChatGPT suggested that the player should have a chance to increase their energy restoration rate, decrease their energy consumption, or increase the invisibility time duration as an alternate reward. 

Two suggested features from ChatGPT\textemdash a shortcut algorithm which would allow the player to take alternate paths from the generated rooms and an enhanced enemy AI patrolling feature\textemdash could not be implemented due to bugs and a context misunderstanding in the code provided by ChatGPT. Finally, ChatGPT suggested to add game feel elements that included character animations (implemented in the form of slight idle movement), a UI indicator for when the player became invisible, color changing effects on the enemy for both when they detected the player and on the player for when they were low on health, and proximity-based sound effects when enemies moved on the map.



\section{Experiment Design and User Study}

\subsection{Website}

A website was hosted on Github Pages to serve as a hub for users to play the games. 
The games were anonymized and the ordering of the base, human-designed, and ChatGPT-suggested games were shuffled so they could be evaluated by participants for quality, feel, and gameplay. Users would select a genre of game and then be given the anonymized games (labeled A-C) to play in order. This experiment setup was inspired by Mazzone's art and AI collaborative output experiments, which asked participants to evaluate AI generated art in a Turing Test style experiment \cite{mazzone2019art}. The games' pages included instructions for how to play and control them below the game window as well as arrows to navigate between games. 
The participants could play the games for as long as they liked. After 2 minutes, a button would flash orange on screen\textemdash either the `next' button for the user to play the next game in the genre or the button to the user study form\textemdash to encourage users to evaluate all of the games in the study.

\subsection{Survey}

We sent out the study as a Google Form survey via Twitter posts, flyers (one of which was featured earlier in this paper as Figure \ref{fig:poster} as a co-designed ChatGPT-human creation), and Discord servers. For the first section of the form, we asked general demographic questions concerning the gaming habits, their game design experience, and experience with AI. The demographic survey questions are listed in Table \ref{tab:demo-q}. The following sections of the survey were divided based on the three genres of games (space shooter, platformer, and roguelike). The participants did not have to play all 3 genres, but were required to play all 3 games in the genre of their choice. After they played the games on the hub website, participants were asked to rank the games (described as they were on the website as game A, B, and C for anonymity) based on their aesthetic and qualitative evaluations. Finally, the participants were asked to give optional feedback on the games they played. We received a total of 45 responses.

\begin{table}[h]
\centering
\begin{tabular}{|p{0.05\linewidth}|p{0.85\linewidth}|}
\rowcolor[HTML]{C0C0C0} 
\hline
Q\# & Question                                                           \\ \hline
Q1  & On average, how often do you play video games?                     \\ \hline
Q2  & What are some of your favorite genres of video games to play?      \\ \hline
Q3 & What type of games do you play more? Independently made games or triple-A games?                   \\ \hline
Q4  & What is your level of game design experience?                      \\ \hline
Q5  & What kinds of games do you design?                                 \\ \hline
Q6 & Have you ever submitted a game or prototype to a game jam (either leisurely or for an assignment)? \\ \hline
Q7  & Have you ever used an AI-assisted content generation tool?         \\ \hline
Q8  & On average, how often do you interact with a large language model? \\ 
\hline
\end{tabular}
\vspace{3px}
\caption{\textbf{Demographic survey questions asked in the user study separated by general topic.}}
\label{tab:demo-q}
\end{table}

\section{User Study Results}
\subsection{Demographic Information}

All of the participants in the study played video games (2 or more hours a week) with 48\% of participants playing 8 or more hours on average a week. We asked participants about the types of games they played in the context of publishers: independent (i.e. games published published on user-based platforms such as Newgrounds or Itch.io and released by small indie studios such as Supergiant, Innersloth, Developer Digital, etc.) and/or triple-A games (i.e. games published largely on console systems and released by large game studios such as Nintendo, FromSoftware, Riot, etc.) 57\% of participants reported to equally play both independent and triple-A titles and 31\% played independent games more frequently. 53\% of the participants had never designed a game, while 31\% had between 1-11+ years of experience making games.  42\% of those who had experience with game design had submitted a game or prototype to a game jam, while 24\% reported an attempt to submit to a jam but had never finished a game for it. 64\% of participants had experience using and interacting with an AI-assisted content generation tool (e.g. text-to-image generators, LLMs). 62\% of participants reported interacting with a large language model between 1 to 4 hours a week.

We asked participants to list the genres of games they like to play and, if they designed games, to list the genres they like to design. 
Platformer games were tied with puzzle games for the most commonly made game genre and the 5th most played genre. Roguelikes were the 4th most frequently made genre and the 11th most played genre. While arcade games were not explicitly said as a genre, we could consider the space shooter game an action game. Action games were the 2nd most played game and the 2nd to least most designed game. From this, we can gather that players were most familiar with action games, while game designers were more experienced with designing platforming games. Roguelikes would not be as familiar to players or designers. 

\subsection{Game Ranking}


\input{ext_tex/color_rank_table}


For each game genre, we asked participants to rank the three anonymized games from ``Best'', ``Mid'', and ``Worst'' based on the following 6 categoies of criteria: \textit{$C_{1}$, overall preference}; \textit{$C_{2}$, game feel}; \textit{$C_{3}$, innovation}; \textit{$C_{4}$, thematic cohesion}; \textit{$C_{5}$, most interesting abilities}; and \textit{$C_{6}$, visual presentation}.
For example, for the ``innovation'' category a participant can rank the games as ``Best'' = A, ``Mid'' = C, and ``Worst'' = B for a given genre. 
Participants were not required to play all 3 genres of games, and some participants chose not to answer every question, therefore each question has a varying number of responses. Our experimental hypothesis was that the human-designed game would outrank the ChatGPT-guided game while both would outrank the baseline game.


To examine the difference in votes across the 3 games for each category, we created a scoring system based on the following equation $S_{G} = 3V_{B} + 2V_{M} + V_{W}$
where $V_{B}$ is the number of votes the game received for being ``Best'' in the category, $V_{M}$ is the number of votes the game received for being ``Mid'' in the category, and $V_{W}$ is the number of votes the game received for being the ``Worst'' of the category. The ``Best'', ``Mid'', and ``Worst'' votes are weighted with values of 3,2,1 respectively and summed to equal the final score $S_{G}$ for a given game $G$ where $G \in $ [Human game, ChatGPT game, Base game]. Table \ref{tab:score-table} shows the scores received in the category out of all of the votes (each category received different numbers of total votes). The platformer genre had the most variance in ranking, with 2 categories (game feel and visual presentation) going against the hypothesis. From this, there were potentially some evaluative disagreement and mixed opinion for comparatively evaluating the quality of the platformer games. The space shooter genre only had one category deviate (innovation). The roguelike category contained no scores that differed from the hypothesis. 
Overall, participants almost consistently rated the human-designed games higher than the ChatGPT-suggested games, and both games were better than the control baseline game with no added game features.

\subsection{Participant Feedback and Comments}

While optional, we received a total of 22, 20, and 19 comments for the space shooter, platformer, and roguelike games respectively.
55\% of the comments left were from participants that reported to have some experience designing games. We performed simple word frequency analysis on the comments using SpaCy\footnote{https://spacy.io/} to look for commonality in themes and criticisms concerning the game. Some critiques were directed towards individual games while others critiqued the games at a general level. For readability, we refer to the games the users critiqued as their true labels for the following paragraphs.

For the space shooter games, 91\% comments left by participants concerned the player ``ability'', while 70\% of the comments pertained to the ``feel'' of the game. Participants commented positively on the ``cool animations'' and ``interesting''-ness of the abilities of the game. Many participants even offered suggestions for improvement on the abilities such as adding more enemies and changing the rate of fire. However, some participants had issues with the controls and accessibility of the game. Other found some of the sound effects ``harsh'' and the overall gameplay ``easy'' or ``boring'' and the abilities ``unnecessary''. Some participants had positive comments about the ChatGPT-suggested game.

For the platformer games, all of the comments concerned the ``feel'' of the game and half concerned the ``level''. Unfortunately, a majority of the comments were negative. Some participants stated the controls were too ``confusing'' and ``horrendous'' -- particularly in the ChatGPT-suggested and the human-designed platformer games. Although they found the abilities and mechanics ``interesting'' and ``floaty'', many people stated that the games were ``not polished''. Some participants said that the base game, while many participants stated had better controls, had ``boring'' level design. One participant even suspected that ``the games may have been by an AI'' but could not identify which. 

Lastly, the roguelike games were more varied in responses and had only 26\% comments pertaining to the feel of the game and 21\% related to the ``time'' of the game. Many of the comments related to the ``feedback'' given by the game to players and the general ``combat.'' However, there were some complaints about ``broken levels'' in the games, a possible generation error from the PCG levels. Many participants explicitly stated that they most enjoyed the free-movement of the shuriken ability from the human-designed game. Many participants stated that the movement of the players and enemies felt ``stiff'' and commented their ``confusion'' for the gameplay and saying the game was ``unclear''.


\section{Discussion}

\subsection{Human-ChatGPT Game Genre and Game Engine Knowledge Base}

As the game developers and designers for the study, the overall quality of the games were naturally limited by our skill levels. ChatGPT, as a general large-language model, can only hallucinate ``reasonable'' game design decisions and regurgitate relevant code snippets. The Unity3D engine has a large breadth of online resources available to developers through forum sites, in addition to online tutorials and open source code projects that provide code examples. With this information, ChatGPT can suggest code that is relevant to the task at hand and contextual enough for the Unity engine. Had this experiment been done in an engine with fewer online resources or one which does not rely on text-based programming such as Scratch, ChatGPT's suggestions may not have been as helpful. 

Understanding of common tropes, important gameplay elements, and basic problem solving was also an important factor in the human-designed games which separated them from their AI counterparts. Inferring the need for ``coyote jumping'' in the human designed platformer simply because the jump ``felt off'' was not a service ChatGPT could provide. In situations where problems would arise in the code implementation, ChatGPT would often struggle with suggesting fixes without pointed prompting from the designer, which required a deeper understanding of programming and certain aspects of Unity. For instance, when recreating the base game of the space shooter, ChatGPT provided the correct method for destroying enemies on collision with the player, but failed to call the method at any point in the script. While the prompter immediately recognized the issue and how to fix it, ChatGPT was unable to identify the problem itself despite being given numerous descriptions of the error. 

In many instances, the prompter was forced to incorporate elements against their better judgment in an attempt to role-play as an individual who lacked any design instincts. In some cases, those instincts made their way into the game design without the prompter realizing. Enemy spawn points in the AI-recreated shooter were instinctively placed off-screen so they would not spawn directly into the middle of the scene. This was not suggested by ChatGPT, but was a knee-jerk impulse from the prompter. Divorcing the game making process from the designer’s instincts entirely to force sole-dependence on ChatGPT's output proved challenging in itself.

ChatGPT performed best when it was prompted for abstract, high level game feature suggestions. These feature suggestions were genre relevant and specific, occasionally matching the features implemented for the human-designed games. Time manipulation in the space shooter, size manipulation in the platformer, and invisibility in the roguelike were all present in both versions of each game. These suggestions were helpful and sometimes aligned with our creative visions for the game. However, to maintain consistency, we had to rely on ChatGPT's code interpretation and attempt to implement it directly into the Unity engine. 

Conversely, ChatGPT struggled with context, particularly when suggesting code snippets to implement into the game. For the roguelike game, it would not account for the turn-based movement of the game and would recommend real-time movement and actions instead. It also struggled with more complex code concepts such as procedural content generation and enemy AI movement. For the shooter and platformer games, it struggled with Unity’s coordinate system, passing incorrect values for rotation and position and sometimes mixing code for 2D and 3D games. Therefore, we would recommend ChatGPT as a tool for conceptual game ideas or for writing pseudocode over direct code implementation in a game engine.

\subsection{Design Approach}
The initial design approach from the human designer included the use of multiple online resources. This included instructions on how to create particle effects and toggle camera shake. While this may bring into question the designers usefulness\textemdash even perhaps blurring the lines between this approach and one lead by AI\textemdash an important distinction comes in the designer’s ability to pursue the right ideas, and their proclivity to remix existing ideas which better serve their creative vision. For instance, the glitch effect indicating damage taken in the human-designed shooter was the result of a multi-step process: scouring online for chromatic aberration effects; happening upon a tutorial for glitching UI elements; and implementing a modified version of the glitch to cover the screen when the player was hit. 
This effect also influenced the decision to change the color palette of the game each time the player was hit, almost as if the player was shifting to a different dimension when attacked. The process began with searching for a basic color shifting script, but transformed into something else entirely which better suited the designer’s goals. 
This made for the early stages to be less of an exercise in creating games elements from scratch, and rather in curating them in an appropriate way. The LLM, by contrast, would remain within the confines of what was asked of it, and could not replicate any semblance of the same vision.

All three human-designed games notably displayed similar patterns during ideation. The designer would often produce an output which took context into account, maintaining a consistent theme and tying each element together with complimenting visual effects. Using splash-screen color effects when activating each ability in the shooter, for instance, or developing abilities which would reasonably fall inside the definition of “shapeshifter” for the platformer. However, they were also often susceptible to a narrow outlook following the initial brainstorming stage. In the case of the shooter, every ability was a screen-wide offensive weapon and no real consideration was given to defensive tools or ones that affected movement. Essentially, while the designer would often attempt to initially “think outside the box”, once they committed to an idea they would then continue along a similar line of thinking for the remainder of the ideation process. These cognitive biases may be better defined as fixed functionality [reference] or the Einstellung effect [reference], where an individual will work within a smaller, standardized range of options despite the breadth of alternatives which are available. While this may certainly be in part the result of the designer’s experience level, this consistently observable behavior presents an interesting dichotomy between the human and AI approach.

The LLM offered a much wider range of suggestions without falling victim to the same pattern. The user study indicates that although the human platformer demonstrated much stronger thematic cohesion, one of the more interesting ideas- gravity manipulation- came from the AI platformer. Depending on the needs of the project, either metric may be more relevant than the other, but having access to both would certainly allow the developer to make the most informed decision.

\subsection{ChatGPT as an Ink Splotch}
The exercise of analyzing structureless blotches of ink and extracting meaning or images from them is common in multiple fields. In psychology it takes the form of the Rorschach test, where patients identify the images they see within the ink blotches. This is then analyzed to understand their pattern of thinking. The artistic process of Bulletism, where artists will splash a canvas with ink and use whichever images they see as inspiration for their next work\cite{Artsapien_2023}, is commonly practiced by fledgling artists or those who have reached a creative impasse. In separating the human-driven and AI-driven approaches to game design and then removing the barrier between the two, it became apparent that LLMs such as ChatGPT could fit a similar role for game designers. 

In each genre, the user study indicates enough votes for ChatGPT games in the “most interesting ideas” field to suggest that LLMs can suggest compelling enough ideas to explore further. ChatGPT’s ability to provide inspiring design choices was also noted by the human designers themselves after reading its suggestions. Notably, the human designer became aware of their over-focused design process for the top-down shooter when reading AI suggestions such as:

\begin{itemize}
    \item \textbf{Energy Shield}: Activate a protective shield that absorbs a certain amount of damage (a defensive ability).
    \item \textbf{Afterburner boost}: A short burst of speed that allows the player to quickly evade enemy fire or close the gap between them and their targets (a movement ability).
    \item \textbf{Repair drones}: Release drones that repair your ship or restore its shields over time (a passive ability).
\end{itemize}

The human designer acknowledged they had not considered non-offensive options for abilities, and nearly instantly began to absorb these suggestions into their “elemental” theme for the game. The energy shield suggestion was reimagined as a protective barrier of rocks and dirt, and the afterburner as an expulsion of dark matter. The human designer continued to curate the game and understand the importance of theme and consistency, while the AI provided an uninhibited list of recommendations to jumpstart that creative process. 


\section{Future Work}
The crux of this experiment was to synthesize a situation which would not often occur outside of the context of research. This was done to isolate the procedures and explore the merits of each without cross contamination. In the future, we would like to tackle experiments which better recreate real-life scenarios of human-AI creative collaborations and how they impact both the developers and the audience. One such study could involve instructing experienced developers to use LLMs in a similar fashion for creative inspiration, but permitting full control over which suggestions they would take and how they would implement them. This would then be recreated multiple times by different developers with the same prompt, with developer feedback indicating the usefulness of the process and the end products being examined for homogeneity. It may also be pertinent to test multiple different LLMs on the same premise, and expand the creative liberties of the models by inciting their help for level design, encounter design, and a more diverse range of gameplay features, particularly in different game engines with less online resources such as Godot, CryEngine, and even private, proprietary engines. 

\section{Conclusion}
In developing a minimalist game focused on unique game mechanics and maximizing game feel, then reproducing the process using an LLM, the resounding response from a panel of unbiased, blind judges was that the AI driven games could not compare to those made by humans. In all but one metric that was measured by the survey, the human designed games\textemdash which had been given the same restrictions and general time constraints as their AI counterparts\textemdash ranked the highest. While ChatGPT was able to produce fully satisfactory and structurally sound games in a leading role, it had no direct ability to replace a human designer’s creative goals, understanding of important context, and ability to identify what “felt” good to a player. These findings reject the notion that AI could act as a sufficient replacement for humans in game design, despite their ability to be highly effective tools in that process. However, certain observations from each game- including their ability to innovate and employ interesting gameplay features- indicated that the AI games had enough appeal for players to want to see them explored in full. These results, compounded by the human designer’s own acknowledgment of their shortcomings in ideation and the expanded outlook they gained through exploring ChatGPT’s suggestions, indicate that AI can have an effective role in the creative process of game making without undercutting the work of game developers. 


\bibliographystyle{ACM-Reference-Format}
\bibliography{bibliography}


\appendix
\newpage

\begin{figure*}[h!]
\centering
    \begin{subfigure}[]{0.45\textwidth}
        \centering
        \includegraphics[width=0.9\linewidth]{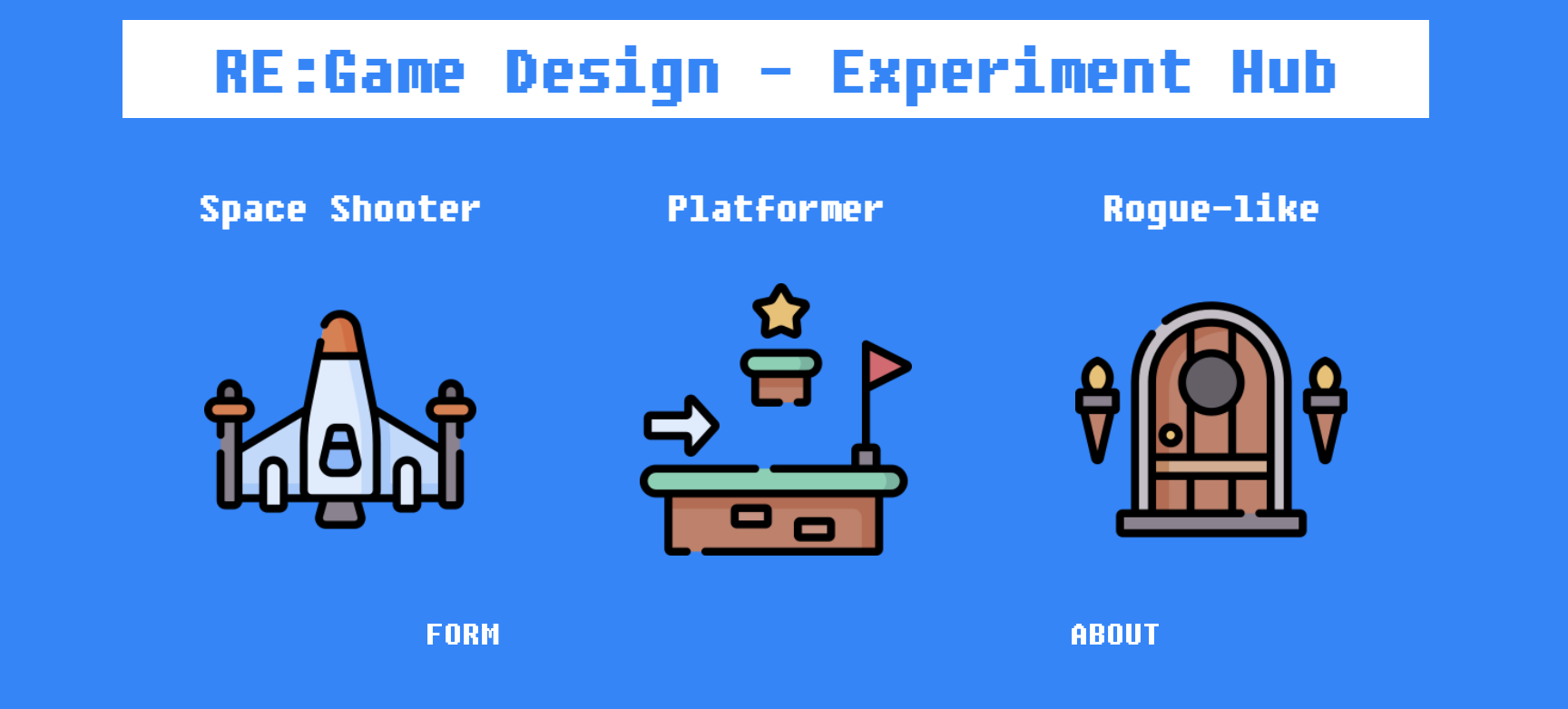}
        \caption{Screen-capture of the website where users could play each of the 9 games and provide feedback.}
        \label{fig:website}
    \end{subfigure}
    ~ 
    \begin{subfigure}[]{0.55\textwidth}
        \centering
    \includegraphics[width=0.9\linewidth]{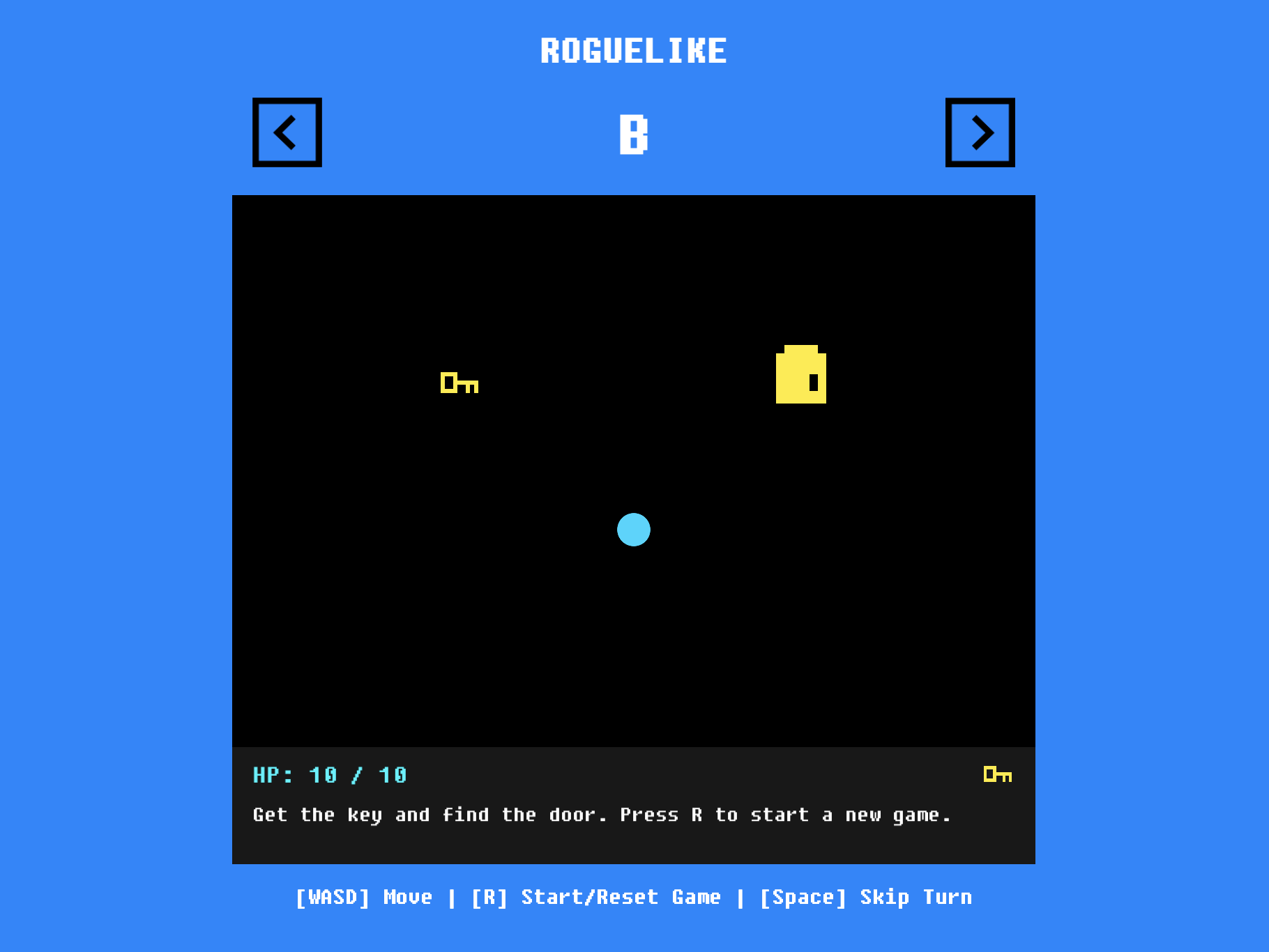}
    \caption{The base roguelike game labeled as `B' on the website. Instructions and links to the form and home page are located below the game.}
    \label{fig:platformer-c}
    \end{subfigure}
\end{figure*}



\begin{figure*}[h!]
    \centering
    \begin{subfigure}[t]{0.5\textwidth}
        \centering
        \includegraphics[width=0.95\textwidth]{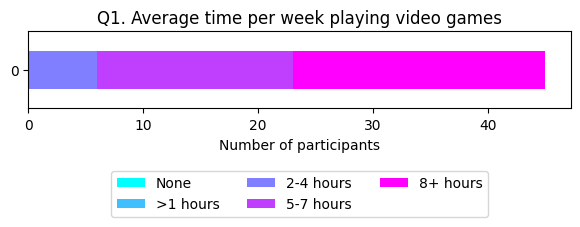}
        \caption{Average time per week participants played video games (Q1)}
        \label{fig:game-play}
    \end{subfigure}%
    ~ 
    \begin{subfigure}[t]{0.5\textwidth}
        \centering
        \includegraphics[width=0.95\textwidth]{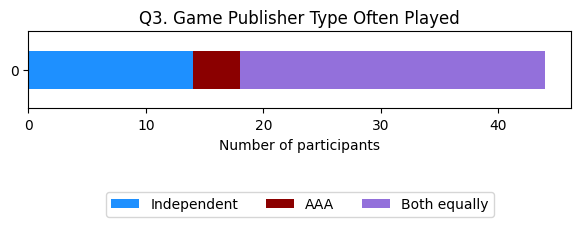}
        \caption{Report of the types of games often played by participants (Q3)}
        \label{fig:game-design-exp}
    \end{subfigure}
\end{figure*}


\begin{figure*}[h!]
    \centering
    \begin{subfigure}[t]{0.5\textwidth}
        \centering
        \includegraphics[width=0.95\textwidth]{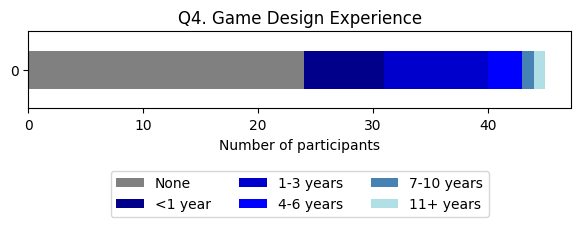}
        \caption{Level of game design experience of participants by years (Q4)}
        \label{fig:game-design-exp}
    \end{subfigure}%
    ~ 
    \begin{subfigure}[t]{0.5\textwidth}
        \centering
        \includegraphics[width=0.95\textwidth]{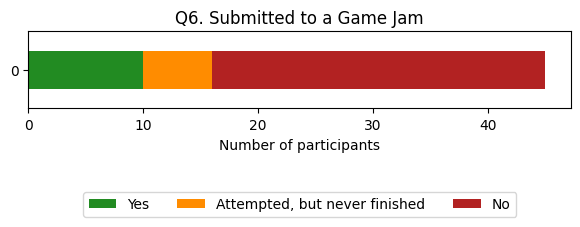}
        \caption{Experience submitting a prototype game to a game jam (Q6)}
        \label{fig:game-jam-exp}
    \end{subfigure}
\end{figure*}


\begin{figure*}[h!]
    \centering
    \begin{subfigure}[t]{0.5\textwidth}
        \centering
        \includegraphics[width=0.95\textwidth]{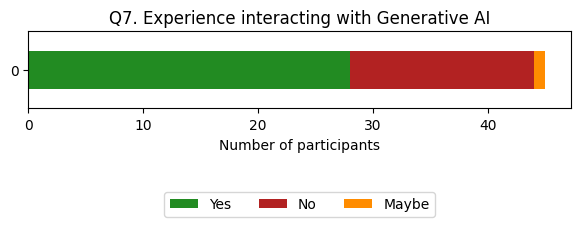}
        \caption{Level of game design experience of participants by years (Q7)}
        \label{fig:}
    \end{subfigure}%
    ~ 
    \begin{subfigure}[t]{0.5\textwidth}
        \centering
        \includegraphics[width=0.95\textwidth]{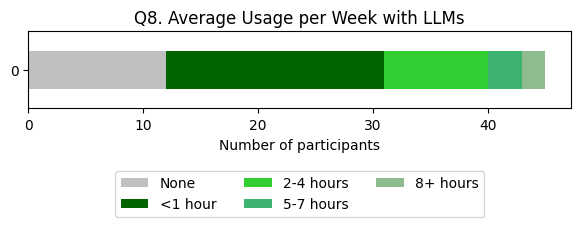}
        \caption{Experience interacting with LLMs averaged on a weekly basis (Q8)}
        \label{fig:llm-exp}
    \end{subfigure}
\end{figure*}

\begin{figure*}[h!]
    \centering
    \includegraphics[width=0.8\linewidth]{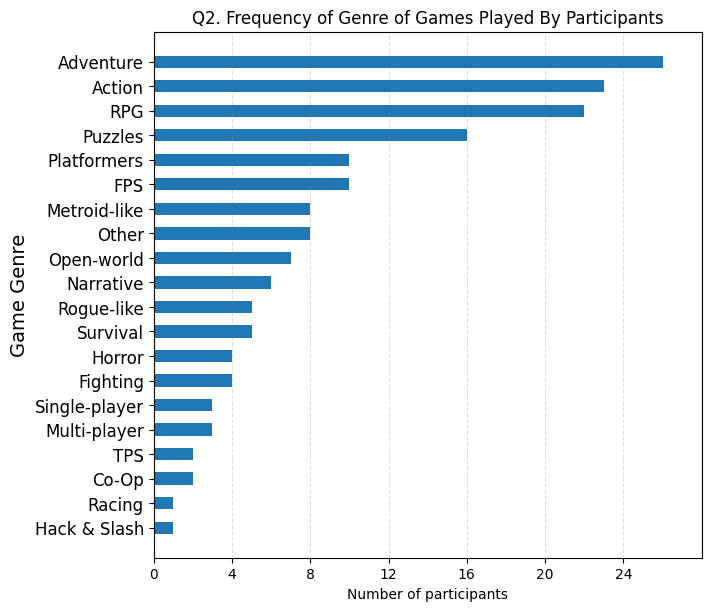}
    \caption{\textbf{Frequency of preferential genres of games to play from all participants.}}
    \label{fig:game-play-freq}
\end{figure*}

\begin{figure*}[h!]
 \centering
        \includegraphics[width=0.8\linewidth]{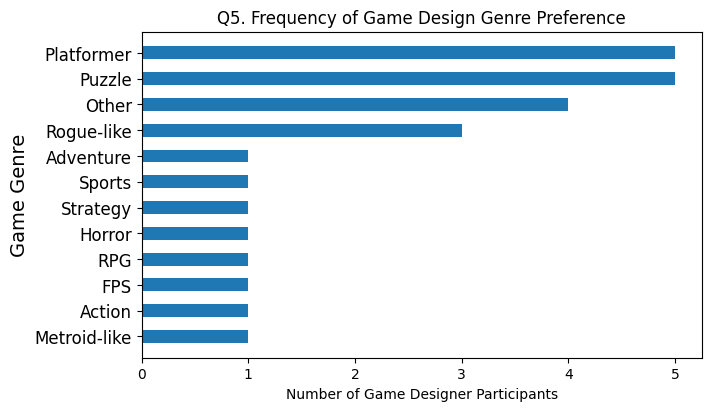}
        \caption{\textbf{Frequency of reported genres designed from game designer participants.}}
        \label{fig:game-design-freq}
\end{figure*}

\begin{table*}[h!]
\begin{tabular}{|lccc|}
\hline
\multicolumn{4}{|c|}{\cellcolor[HTML]{CBCEFB}Space Shooter}                                                    \\ \hline
\multicolumn{1}{|l|}{} &
  \multicolumn{1}{l|}{\% {[}Best = Human{]} Vote} &
  \multicolumn{1}{l|}{\% {[}Mid = ChatGPT{]} Vote} &
  \multicolumn{1}{l|}{\% {[}Worst = Base{]} Vote} \\ \hline
\multicolumn{1}{|l|}{Overall preference}        & \multicolumn{1}{c|}{70.7} & \multicolumn{1}{c|}{70.7} & 85.4 \\
\multicolumn{1}{|l|}{Game feel}                 & \multicolumn{1}{c|}{77.5} & \multicolumn{1}{c|}{72.5} & 87.5 \\
\multicolumn{1}{|l|}{Innovation}                & \multicolumn{1}{c|}{57.5} & \multicolumn{1}{c|}{66.7} & 87.2 \\
\multicolumn{1}{|l|}{Thematic Cohesion}         & \multicolumn{1}{c|}{53.8} & \multicolumn{1}{c|}{\cellcolor[HTML]{C0C0C0}35.9} & 56.4 \\
\multicolumn{1}{|l|}{Interesting Game Features} & \multicolumn{1}{c|}{62.5} & \multicolumn{1}{c|}{60}   & 90   \\
\multicolumn{1}{|l|}{Visual Presentation}       & \multicolumn{1}{c|}{77.5} & \multicolumn{1}{c|}{72.5} & 90   \\ \hline
\end{tabular}
\caption{\textbf{Aggregated percentage of votes split for the space shooter genre.} Cells with less than half of the majority vote and that go against the hypothesis are colored in gray.}
\label{tab:space-shooter-votes}     
\end{table*}

\begin{table*}[h!]
\begin{tabular}{|lccc|}
\hline
\multicolumn{4}{|c|}{\cellcolor[HTML]{AAE4A9}Platformer}                                                                                                                                               \\ \hline
\multicolumn{1}{|l|}{}                          & \multicolumn{1}{l|}{\% {[}Best = Human{]} Vote} & \multicolumn{1}{l|}{\% {[}Mid = ChatGPT{]} Vote} & \multicolumn{1}{l|}{\% {[}Worst = Base{]} Vote} \\ \hline
\multicolumn{1}{|l|}{Overall preference}        & \multicolumn{1}{c|}{64.1}                       & \multicolumn{1}{c|}{\cellcolor[HTML]{C0C0C0}47.4}                        & 52.6                                            \\
\multicolumn{1}{|l|}{Game feel}                 & \multicolumn{1}{c|}{\cellcolor[HTML]{C0C0C0}48.7}                       & \multicolumn{1}{c|}{\cellcolor[HTML]{C0C0C0}48.7}                        & \cellcolor[HTML]{C0C0C0}33.3                                            \\
\multicolumn{1}{|l|}{Innovation}                & \multicolumn{1}{c|}{71.1}                       & \multicolumn{1}{c|}{55.3}                        & 78.9                                            \\
\multicolumn{1}{|l|}{Thematic Cohesion}         & \multicolumn{1}{c|}{64.9}                       & \multicolumn{1}{c|}{\cellcolor[HTML]{C0C0C0}40.5}                        & 51.4                                            \\
\multicolumn{1}{|l|}{Interesting Game Features} & \multicolumn{1}{c|}{59}                         & \multicolumn{1}{c|}{\cellcolor[HTML]{C0C0C0}43.6}                        & 87.2                                            \\
\multicolumn{1}{|l|}{Visual Presentation}       & \multicolumn{1}{c|}{67.6}                       & \multicolumn{1}{c|}{\cellcolor[HTML]{C0C0C0}32.4}                        & 32.4                                            \\ \hline
\end{tabular}
\caption{\textbf{Aggregated percentage of votes split for the platformer genre.} Cells with less than half of the majority vote and that go against the hypothesis are colored in gray.}
\label{tab:platformer-votes}
\end{table*}

\begin{table*}[h!]
\begin{tabular}{|lccc|}
\hline
\multicolumn{4}{|c|}{\cellcolor[HTML]{E4A9A9}Roguelike}                                                                                                                                                \\ \hline
\multicolumn{1}{|l|}{}                          & \multicolumn{1}{l|}{\% {[}Best = Human{]} Vote} & \multicolumn{1}{l|}{\% {[}Mid = ChatGPT{]} Vote} & \multicolumn{1}{l|}{\% {[}Worst = Base{]} Vote} \\ \hline
\multicolumn{1}{|l|}{Overall preference}        & \multicolumn{1}{c|}{71.4}                       & \multicolumn{1}{c|}{50}                          & 60.5                                            \\
\multicolumn{1}{|l|}{Game feel}                 & \multicolumn{1}{c|}{63.2}                       & \multicolumn{1}{c|}{65.8}                        & 63.2                                            \\
\multicolumn{1}{|l|}{Innovation}                & \multicolumn{1}{c|}{80}                         & \multicolumn{1}{c|}{71.4}                        & 74.3                                            \\
\multicolumn{1}{|l|}{Thematic Cohesion}         & \multicolumn{1}{c|}{57.1}                       & \multicolumn{1}{c|}{68}                          & 51.4                                            \\
\multicolumn{1}{|l|}{Interesting Game Features} & \multicolumn{1}{c|}{83.3}                       & \multicolumn{1}{c|}{72.2}                        & 69.4                                            \\
\multicolumn{1}{|l|}{Visual Presentation}       & \multicolumn{1}{c|}{63.9}                       & \multicolumn{1}{c|}{61.1}                        & 61.1                                            \\ \hline
\end{tabular}
\caption{\textbf{Aggregated percentage of votes split for the roguelike genre.} Cells with less than half of the majority vote and that go against the hypothesis are colored in gray.}
\label{tab:roguelike-votes}
\end{table*}

\end{document}

%% file: ext_tex/color_rank_table.tex
\begin{table*}[]
\begin{tabular}{l|rrr|rrr|rrr|}
\cline{2-10}
 &
  \multicolumn{3}{c|}{\cellcolor[HTML]{CBCEFB}Space Shooter} &
  \multicolumn{3}{c|}{\cellcolor[HTML]{FFCE93}Platformer} &
  \multicolumn{3}{c|}{\cellcolor[HTML]{C0C0C0}Roguelike (\%)} \\ \cline{2-10} 
 &
  \multicolumn{1}{l}{Human} &
  \multicolumn{1}{l}{ChatGPT} &
  \multicolumn{1}{l|}{Base} &
  \multicolumn{1}{l}{Human} &
  \multicolumn{1}{l}{ChatGPT} &
  \multicolumn{1}{l|}{Base} &
  \multicolumn{1}{l}{Human} &
  \multicolumn{1}{l}{ChatGPT} &
  \multicolumn{1}{l|}{Base} \\ \hline
\multicolumn{1}{|l|}{Overall Preference} &
  \cellcolor[HTML]{CCEDCC}110 &
  \cellcolor[HTML]{FAF8C4}88 &
  \cellcolor[HTML]{E4A9A9}48 &
  \cellcolor[HTML]{CCEDCC}97 &
  \cellcolor[HTML]{FAF8C4}77 &
  \cellcolor[HTML]{E4A9A9}57 &
  \cellcolor[HTML]{CCEDCC}87 &
  \cellcolor[HTML]{FAF8C4}77 &
  \cellcolor[HTML]{E4A9A9}64 \\
\multicolumn{1}{|l|}{Game Feel} &
  \cellcolor[HTML]{CCEDCC}108 &
  \cellcolor[HTML]{FAF8C4}85 &
  \cellcolor[HTML]{E4A9A9}48 &
  \cellcolor[HTML]{CCEDCC}86 &
  \cellcolor[HTML]{E4A9A9}73 &
  \cellcolor[HTML]{FAF8C4}75 &
  \cellcolor[HTML]{CCEDCC}92 &
  \cellcolor[HTML]{FAF8C4}71 &
  \cellcolor[HTML]{E4A9A9}65 \\
\multicolumn{1}{|l|}{Innovation} &
  \cellcolor[HTML]{FAF8C4}93 &
  \cellcolor[HTML]{CCEDCC}95 &
  \cellcolor[HTML]{E4A9A9}49 &
  \cellcolor[HTML]{CCEDCC}104 &
  \cellcolor[HTML]{FAF8C4}76 &
  \cellcolor[HTML]{E4A9A9}48 &
  \cellcolor[HTML]{CCEDCC}93 &
  \cellcolor[HTML]{FAF8C4}65 &
  \cellcolor[HTML]{E4A9A9}52 \\
\multicolumn{1}{|l|}{Thematic Cohesion} &
  \cellcolor[HTML]{CCEDCC}91 &
  \cellcolor[HTML]{FAF8C4}80 &
  \cellcolor[HTML]{E4A9A9}63 &
  \cellcolor[HTML]{CCEDCC}98 &
  \cellcolor[HTML]{FAF8C4}66 &
  \cellcolor[HTML]{E4A9A9}58 &
  \cellcolor[HTML]{CCEDCC}79 &
  \cellcolor[HTML]{FAF8C4}66 &
  \cellcolor[HTML]{E4A9A9}65 \\
\multicolumn{1}{|l|}{Most Interesting Abilities} &
  \cellcolor[HTML]{CCEDCC}101 &
  \cellcolor[HTML]{FAF8C4}92 &
  \cellcolor[HTML]{E4A9A9}47 &
  \cellcolor[HTML]{CCEDCC}103 &
  \cellcolor[HTML]{FAF8C4}85 &
  \cellcolor[HTML]{E4A9A9}46 &
  \cellcolor[HTML]{CCEDCC}96 &
  \cellcolor[HTML]{FAF8C4}65 &
  \cellcolor[HTML]{E4A9A9}55 \\
\multicolumn{1}{|l|}{Visual Presentation} &
  \cellcolor[HTML]{CCEDCC}107 &
  \cellcolor[HTML]{FAF8C4}83 &
  \cellcolor[HTML]{E4A9A9}50 &
  \cellcolor[HTML]{CCEDCC}98 &
  \cellcolor[HTML]{E4A9A9}56 &
  \cellcolor[HTML]{FAF8C4}68 &
  \cellcolor[HTML]{CCEDCC}83 &
  \cellcolor[HTML]{FAF8C4}75 &
  \cellcolor[HTML]{E4A9A9}58 \\ \hline
\end{tabular}
\caption{\textbf{Scores for each game and genre using the formula $S_{G}$ and category criteria} Models with the highest score are colored in green, second highest in yellow, and lowest in red.}
\label{tab:score-table}
\end{table*}